\documentclass[letterpaper]{article} 
\usepackage{aaai24}  
\usepackage{times}  
\usepackage{helvet}  
\usepackage{courier}  
\usepackage[hyphens]{url}  
\usepackage{graphicx} 
\usepackage{booktabs} 
\usepackage{natbib}  
\usepackage{caption} 
\usepackage{amsmath}
\usepackage{amssymb}
\usepackage{algorithm}
\usepackage{algorithmic}

\usepackage{newfloat}
\usepackage{listings}
\DeclareCaptionStyle{ruled}{labelfont=normalfont,labelsep=colon,strut=off} 
\floatstyle{ruled}
\newfloat{listing}{tb}{lst}{}
\floatname{listing}{Listing}
\title{Exploring the Optimization Objective of One-Class Classification for Anomaly Detection}
\author {
    Han Gao\textsuperscript{\rm 1,2,3},\\
    Huiyuan Luo\textsuperscript{\rm 1,3} \equalcontrib,
    Fei Shen\textsuperscript{\rm 1,2,3,4},
    Zhengtao Zhang\textsuperscript{\rm 1,2,3,4,5}\thanks{Corresponding author.}
}
\affiliations {
    \textsuperscript{\rm 1}Institute of Automation, Chinese Academy of Sciences, Beijing 100190, People's Republic of China\\
    \textsuperscript{\rm 2}University of Chinese Academy of Sciences, Beijing 100049, People's Republic of China\\
    \textsuperscript{\rm 3}Engineering Laboratory for Intelligent Industrial Vision, Institute of Automation, Chinese Academy of Sciences, Beijing 100190, People's Republic of China\\
    \textsuperscript{\rm 4}CASI Vision Technology CO., LTD., Luoyang 471000, People’s Republic of China\\
    \textsuperscript{\rm 5}Binzhou Institute of Technology, Weigiao-UCAS Science and Technology Park, Binzhou, Shandong 256606, People’s Republic of China\\
    gaohan2022@ia.ac.cn, huiyuan.luo@ia.ac.cn, fei.shen@ia.ac.cn, zhengtao.zhang@ia.ac.cn
}

\usepackage{bibentry}

\begin{document}

\maketitle

\begin{abstract}
One-class classification (OCC) is a longstanding method for anomaly detection. With the powerful representation capability of the pre-trained backbone, OCC methods have witnessed significant performance improvements. Typically, most of these OCC methods employ transfer learning to enhance the discriminative nature of the pre-trained backbone's features, thus achieving remarkable efficacy. While most current approaches emphasize feature transfer strategies, we argue that the optimization objective space within OCC methods could also be an underlying critical factor influencing performance. In this work, we conducted a thorough investigation into the optimization objective of OCC. Through rigorous theoretical analysis and derivation, we unveil a key insights: any space with the suitable norm can serve as an equivalent substitute for the hypersphere center, without relying on the distribution assumption of training samples. Further, we provide guidelines for determining the feasible domain of norms for the OCC optimization objective. This novel insight  sparks a simple and data-agnostic deep one-class classification method. Our method is straightforward, with a single $1 \times 1$ convolutional layer as a trainable projector and any space with suitable norm as the optimization objective. Extensive experiments validate the reliability and efficacy of our findings and the corresponding methodology, resulting in state-of-the-art performance in both one-class classification and industrial vision anomaly detection and segmentation tasks.
\end{abstract}

\section{Introduction}

Anomaly detection, a technique leveraging normal samples to identify outliers without laborious sample annotation, proves particularly effective in domains like industrial vision defect detection, where obtaining labeled samples is challenging. Essentially, existing anomaly detection methods use normal samples to train an estimator that recognizes normal patterns. For test samples, the normal estimator gives anomaly scores based on some metrics (e.g., Euclidean distance \cite{cfa,panda,yang2023memseg}, cosine similarity \cite{hu2021semantic,jang2023n}, likelihood \cite{f:cflow,zolfaghari2022unsupervised}, or other metric). Obviously, finding a discriminative normal estimator is a core part of the anomaly detection task. Normal estimators can be constructed by such means as sub-sampled memory of normal \cite{patchcore}, reconstructed encoder-decoder for normal sample \cite{draem}, one-class classification of normal \cite{reiss2023mean,panda}, etc. 

Among these, one-class classification (OCC) is a longstanding method. Traditional techniques like OC-SVM \cite{scholkopf1999support} and SVDD \cite{tax2004support} have found widespread use. With the advent of deep learning, SVDD based methods have embraced auto-encoders \cite{ruff2018deep} and pre-trained backbones \cite{panda} for enhanced feature representation, resulting in notable successes. Generally, the expression for OCC is $ \|\mathcal{\varphi}(\boldsymbol{x}) - \boldsymbol{c}\|^2 \leq R^2$ to map normal samples into the hypersphere, where $\boldsymbol{x}$ is the normal sample, $\varphi(\cdot)$ is the neural network, $R$ is the radius and $\boldsymbol{c}$ is the optimization objective, i.e., the hypersphere center. Empirically, the optimization objectives $\boldsymbol{c}$ are set as average pre-trained features of normal sample or their variants. However, there are still problems. OCC methods often encounter the pattern collapse issue due to neural network $\varphi(\cdot)$ overconfidence \cite{nguyen2015deep}, where positive and negative sample features collapse to the same point. Moreover, domain bias between large pre-trained and smaller anomaly detection datasets can cause similar feature expressions for both sample types. To address these issues, existing OCC methods mainly focus on feature projection, transfer and optimization strategies \cite{reiss2023mean,panda,ruff2018deep}, yielding compact and distinctive normal estimators while mitigating pattern collapse. 

However, despite impressive results achieved by these methods, they overlook the exploration of the optimization objective $\boldsymbol{c}$ in OCC methods, which is another potentially crucial factor influencing detection performance. Is an average of unoptimized pre-trained features the optimal OCC optimization objective? What defines a good optimization objective? These questions remain unexplored and theoretically unexplained. Thus, this paper delves into the OCC method's optimization objective, specifically the center of the hypersphere, and focuses on the following questions:

\begin{itemize}
    \item What theoretical model should be employed to comprehensively describe the optimization objective of OCC?
    \item What properties should the OCC optimization objective have to better separate positive and negative samples for better accuracy?
    \item Can the use of pre-trained feature averages in previous methods be explained by such theory?
    \item Is there a better strategy for setting the OCC optimization objective beyond the pre-trained average? 
\end{itemize}

In this study, based on the above questions, we deeply explore the impact of the optimization objective of OCC. Through rigorous theoretical analysis and derivation, we draw the following conclusions and contributions:
\begin{itemize}
    \item The optimization objective of OCC can be represented in polar coordinates. Feature norm (radius) and similarity (angular) are the two crucial factors.
    \item The polar coordinate space's ability to discriminate between positive and negative samples hinges on reasonable norm values of OCC optimization objective. Inappropriate norm values can lead to significant degradation of the discrimination performance.
    \item The common practice of using pre-trained feature averages aligns with this mathematical model.
    \item Any space with a suitable norm can serve as the optimization objective for OCC model, independent of data. 
\end{itemize}

Elaborations on theoretical derivations and empirical validations are provided in Sections 3, 4, and 5. Moreover, the fourth discovery prompts us to consider the feasibility of employing a data-independent random space as the optimization objective. Consequently, the necessity for pre-trained feature averages predicated on training samples is obviated, allowing for training without the need for initial dataset accumulation. This groundbreaking insight births a data-agnostic deep one-class classification paradigm, thus laying the foundation for propelling traditional offline OCC methods into the realm of online detection. This methodology employs a single $1\times 1$ convolutional layer as a trainable projector and utilizes spaces with suitable norms as optimization objectives. Extensive experiments validate our findings' authenticity and emphasize our approach's effectiveness, resulting in notable performance advancements across one-class anomaly detection and industrial vision anomaly detection and segmentation tasks.

\section{Related Work}

\textbf{Pre-trained based OCC methods:} Traditional OCC methods like OC-SVM \cite{scholkopf1999support} and SVDD \cite{tax2004support} rely on handcrafted features and kernel functions for mapping features to hyperplanes or hyperspheres. The integration of deep learning has improved feature extraction and mapping. Common deep feature extractors fall into two categories: reconstruction-based networks like Auto-Encoders and pre-trained backbones from large datasets like ImageNet \cite{deng2009imagenet}. However, reconstruction-based networks (e.g., Deep SVDD \cite{ruff2018deep}) often provide compressed, low-resolution, and lossy features, which aren't ideal for anomaly detection.

To enhance feature quality, more OCC methods adopt pre-trained backbones. These methods transfer or adapt pre-trained features to OCC optimization spaces (e.g., averaged pre-trained features). PANDA \cite{panda} employs the continual learning strategy to mitigate pattern collapse issues. MS \cite{reiss2023mean} proposes mean-shifted contrastive loss for better adaptation and provides some exploration of the optimization space of OCC by normalizing pre-trained features (subsequent experiments validate it is a potential solution or instance within our theoretical model). While these methods achieve remarkable results, most lack theoretical exploration of the optimization space, which we believe plays a pivotal role in performance. By exploring this space, we aim to derive theoretical principles, explain the use of pre-trained feature averages, and identify better optimization spaces.


\textbf{Anomaly Detection and Segmentation in Industrial Vision:} Compared to OCC tasks, this task is more of a sensory task, aiming to find samples with spatially distributed abnormal pixels within the same category \cite{rippel2021modeling,jiang2022softpatch}. It requires a finer granularity and better discrimination of positive and negative samples as the distribution of defects is often structured and localized. Patch SVDD \cite{Patchsvdd} introduces the SVDD algorithm at the patch level and employs self-supervised learning to form multiple finer-grained centers, enhancing the model's ability to detect and locate local anomalies. DSPSVDD \cite{zhang2021anomaly} combines deep feature extraction with data structure information, designing an enhanced comprehensive optimization objective by simultaneously minimizing hypersphere volume and network reconstruction error. SE-SVDD \cite{hu2021semantic} proposes a semantic-enhanced anomaly detection method based on Deep SVDD, enhancing feature representation through cosine similarity and estimating pixel-wise anomaly scores using a multi-level structure. While these methods enhance the model's sensitivity to local anomalies and improve OCC performance on this task, existing OCC methods still significantly lag behind the SOTA methods. In comparison, our method achieves SOTA-level results on this task with fewer parameters.

\section{Exploring the Optimization Objective for Anomaly Detection}

In this section, we first reviewed the SVDD and Deep SVDD methods. Based on their mathematical definitions, we conducted a rigorous and comprehensive theoretical analysis and derivation of the optimization objective space in OCC, aiming to find the properties that the optimal optimization objective space should possess. Finally, we designed a simple and effective OCC method.


\subsection{Recall on SVDD and Deep SVDD}\label{sec:re_svdd}
We consider the anomaly detection task where the training set $ \boldsymbol{x}\in \mathcal{X}_{train}$ contains only normal samples, and anomaly samples are only available during the testing phase. Following the notation and theoretical analysis of SVDD, all normal samples need to align with the hypersphere center $\boldsymbol{c}$ of the OCC optimization objective. The process is defined as:
\begin{equation}\label{eqa:f_process}
\fontsize{9pt}{\baselineskip}\selectfont 
F(R,\boldsymbol{c})= R^2
\end{equation}
With the constraints: 
\begin{equation}\label{eqa:process}
\fontsize{9pt}{\baselineskip}\selectfont 
\left\|\boldsymbol{x_i} - \boldsymbol{c}\right\|^2 \leq R^2 , \forall \boldsymbol{x_i}\in \mathcal{X}_{train}
\end{equation}

$R$ is the radius of the sphere. Incorporating Lagrangian, Eq.\ref{eqa:process} can be rewrote:
\begin{equation}\label{eqa:lagrangian}
\fontsize{9pt}{\baselineskip}\selectfont 
L\left(R, \boldsymbol{c},\alpha_{i}\right)=R^{2}-\sum_{i} \alpha_{{i}}\left\{R^{2}-\left\|\boldsymbol{x_i}-\boldsymbol{c}\right\|^{2}\right\}
\end{equation}

$\alpha$ is the Lagrange multipliers. By setting the gradients of the relevant parameters to zero, Eq.\ref{eqa:lagrangian} needs to satisfy the following constraints:
\begin{equation}
\fontsize{9pt}{\baselineskip}\selectfont 
\frac{\partial L}{\partial R}=0 \rightarrow \sum_{i} \alpha_{{i}}=1
\end{equation}
\begin{equation}\label{eqa:c}
\fontsize{9pt}{\baselineskip}\selectfont 
\frac{\partial L}{\partial \boldsymbol{c}}=0 \rightarrow \boldsymbol{c}=\sum_{i} \alpha_{i} \boldsymbol{x_i}
\end{equation}

As a result, the hypersphere center $\boldsymbol{c}$ of SVDD can be represented as a linear combination of training samples. Subsequent works such as Deep SVDD extend Eq.\ref{eqa:c} and represent the hypersphere center $\boldsymbol{c}$ as the average of pre-trained features. In addition to this, Deep SVDD empirically analyzes that improper settings of neural networks can lead to the pattern collapse issue, including all weights of the neural network being zero,inappropriate biases and activation functions. Although Deep SVDD provides valuable guidance, rigorous mathematical analysis is still lacking.

\subsection{Theory Analysis for Optimization Objective}\label{sec:theory_opt_obj}
Previous methods have focused primarily on the design of the network structure, typically by empirically setting the hypersphere center $\boldsymbol{c}$ to the average of the pre-trained features. However, as the optimization objective of the entire network, $\boldsymbol{c}$ is also a potential crucial factor influencing performance. In this section, we will thoroughly explore $\boldsymbol{c}$.

\textbf{Optimization Objective can be data-agnostic.}
Assuming the neural network $\varphi(\cdot)$ is with learnable parameters $W$, Eq.\ref{eqa:lagrangian} is modified as follows:

\begin{equation}\label{eqa:lagrangian_W}
\fontsize{9pt}{\baselineskip}\selectfont 
\begin{aligned}
    {L}\left({R}, \boldsymbol{c}, \alpha_{i}, W\right)&={R}^{2}-\sum_{{i}} \alpha_{i}\left\{{R^2}-\left\|\varphi\left(\boldsymbol{x_i},W\right)-\boldsymbol{c}\right\|^{2}\right\}\\
    =&{R}^{2}-\sum_{{i}} \alpha_{{i}} {R}^{2}+\sum_{{i}} \alpha_{{i}}\left\|\varphi\left(\boldsymbol{x_i},W\right)-\boldsymbol{c}\right\|^{2}  
\end{aligned}
\end{equation}

By setting the gradients of relevant parameters to zero, the constraint in Eq.\ref{eqa:lagrangian_W} needs to be satisfied:
\begin{equation}\label{eqa:r_case}
\fontsize{9pt}{\baselineskip}\selectfont 
\frac{\partial L}{\partial R}=2 R-\sum_{{i}} 2 \alpha_{{i}} {R}=0 \rightarrow \sum_{i} \alpha_{{i}}=1
\end{equation}

\begin{equation}\label{eqa:c_case}
\fontsize{9pt}{\baselineskip}\selectfont 
\begin{aligned}
    \frac{\partial L}{\partial W} & =2 \sum_{{i}} \alpha_{{i}} W\left[\varphi\left(\boldsymbol{x_i},W\right)-\boldsymbol{c}\right]=0 \\
     &\rightarrow \sum_{i} \alpha_{{i}} \varphi\left(\boldsymbol{x_i},W\right)=\boldsymbol{c}, or\,{W=0}
\end{aligned}
\end{equation}

According to Eq.\ref{eqa:c_case} , two situations may occur when the objective function $L$ stops optimization:

\textbf{Case 1:} When all the learnable weights of the neural network $\varphi(\cdot)$ are zero, $W=0$. In this case, the network maps all samples to a single point, resulting in a degenerate solution. This is also pointed out as an inappropriate situation by Deep SVDD. If the neural network $\varphi(\cdot)$ does not have constant parameters, the center $\boldsymbol{c}$ in the optimization becomes a zero constant feature vector. Therefore, we can avoid this situation by limiting $\varphi(\cdot)$ to have only trainable parameters and setting $\boldsymbol{c}$ as a non-zero constant feature vector.

\textbf{Case 2:} For any input sample, $\sum_{i} \alpha_{{i}} \varphi\left(\boldsymbol{x_{i}}\right)=\boldsymbol{c}$. If we fix the hypersphere center $\boldsymbol{c}$ , the neural network, leveraging its powerful non-linear mapping ability, will automatically adjust the learnable parameters $W$ to weight the distribution of the original feature space and map it to the target space. This means that the optimization objective $\boldsymbol{c}$ is decoupled from the data, and can be set arbitrarily without considering the underlying data distribution assumption. This conclusion holds even when generalized to more complex neural networks, as they offer a more powerful non-linear fitting capability to ensure the convergence of the mapping to $\boldsymbol{c}$. In this study, we attempted to set $\boldsymbol{c}$ sampled from standard normal distribution, uniform distribution, all-one distribution, and the average of sample pre-trained features. We observed that the model's anomaly detection performance did not exhibit any degree of deterioration, validating our viewpoint.

However, one necessary condition among these is that all normal features can be mapped to this optimization objective space $\boldsymbol{c}$. If the value of $\boldsymbol{c}$ is set improperly, the normal sample space might not be able to map to the target space, resulting in the model's inability to capture all normal patterns and thus a decrease in accuracy. Meanwhile, the network tends to fit those minority outlier points, leading to pattern collapse. The optimization objective space $\boldsymbol{c}$ should be a subset of the collection of mapped feature spaces. Therefore, as long as the norm of $\boldsymbol{c}$ is reasonably set, the type or distribution of the target space $\boldsymbol{c}$ can be arbitrary and data-independent.

\textbf{What is the Suitable Norm?} When the model approaches convergence, it can be approximated that Eq.\ref{eqa:r_case} and Eq.\ref{eqa:c_case} hold true. Substituting Eq.\ref{eqa:r_case} into Eq.\ref{eqa:lagrangian_W}, we obtain:
\begin{equation}\label{eqa:c_norm}
\fontsize{9pt}{\baselineskip}\selectfont 
\begin{aligned}
    {L}\left(\boldsymbol{c}, \alpha_{{i}}, W\right)
     =\sum_{{i}}\alpha_{{i}} \varphi\left(\boldsymbol{x_{i}},W\right)^{2}-2 \sum_{{i}} \alpha_{{i}} \varphi\left(\boldsymbol{x_{i}},W\right) \boldsymbol{{c}}+\boldsymbol{{c}}^{2} 
\end{aligned}
\end{equation}

With Eq.\ref{eqa:c_case}, the above Eq.\ref{eqa:c_norm} can be simplified:
\begin{equation}
\fontsize{9pt}{\baselineskip}\selectfont 
\begin{aligned}
{L}\left(\alpha_{{i}}, W\right) &=\sum_{{i}} \alpha_{{i}} \varphi\left(\boldsymbol{x_{i}},W\right)^{2}-\left[\sum_{{i}} \alpha_{{i}} \varphi\left(\boldsymbol{x_{i}},W\right)\right]^{2} \\
     &=\sum_{{i}} \alpha_{{i}}\left(\varphi\left(\boldsymbol{x_{i}},W\right), \varphi\left(\boldsymbol{x_{i}},W\right)\right)\\
     &-\sum_{{i}, {j}} \alpha_{{i}} \alpha_{{j}}\left(\varphi\left(\boldsymbol{x_{i}},W\right), \varphi\left(\boldsymbol{x_{j}},W\right)\right)
\end{aligned}
\end{equation}

Let $\boldsymbol{z_{i}}=\varphi\left(\boldsymbol{x_{i}},W\right)$, continuing the simplification:
\begin{equation}
    \fontsize{9pt}{\baselineskip}\selectfont 
    \begin{aligned}
    {L}\left(\alpha_{{i}}, \boldsymbol{z_{i}}\right)&=\sum_{{i}} \alpha_{{i}}\left(\boldsymbol{z_{i}}, \boldsymbol{z_{i}}\right)-\sum_{{i}, {j}} \alpha_{{i}} \alpha_{{j}}\left(\boldsymbol{z_{i}}, \boldsymbol{z_{j}}\right)\\
    &=\sum_{{i}} \alpha_{{i}}\left\|\boldsymbol{z_{i}}\right\|^{2}-\sum_{{i}, {j}} \alpha_{{i}} \alpha_{{j}}\left\|\boldsymbol{z_{i}}\right\| \cdot\left\|\boldsymbol{z_{j}}\right\| \cdot \cos \theta_{{ij}}
    \end{aligned}
\end{equation}

For a training set composed solely of normal samples, we can make the assumption that the norm difference of $\forall {i}, {j}, \boldsymbol{z_{i}} ,\boldsymbol{z_{j}}$ is not significant, and we can approximate it as  $\left\|\boldsymbol{z_{i}}\right\| \approx {r},\left\|\boldsymbol{{z}_{{j}}}\right\| \approx {r}, \cos \theta_{{ij}} \approx 1, \sin \theta_{{ij}} \approx 0$,  thus:

\begin{equation}
\fontsize{9pt}{\baselineskip}\selectfont 
{L}\left(r, \theta, \alpha_{{i}}\right) \approx \sum_{{i}} \alpha_{{i}} r^{2}-\sum_{{i}, {j}} \alpha_{{i}} \alpha_{{j}} r^{2} \cos \theta_{{ij}}
\end{equation}

Taking the derivative of ${L}$ and continuing to optimize the model until it fully converges:
\begin{equation}\label{eqa:L_r}
\fontsize{9pt}{\baselineskip}\selectfont 
\frac{\partial L}{\partial r} \approx 2 r\left(\sum_{i} \alpha_{{i}}-\sum_{{i}, {j}} \alpha_{{i}} \alpha_{{j}}\right)
\end{equation}

\begin{equation}\label{eqa:L_theta}
\fontsize{9pt}{\baselineskip}\selectfont 
\frac{\partial L}{\partial \theta} \approx \sum_{{i}, {j}} \alpha_{{i}} \alpha_{{j}} {r}^{2} \sin \theta_{{ij}}
\end{equation}

With Eq.\ref{eqa:L_r} and $\sin \theta_{{ij}} \approx 0 $, the gradenit of feature norm $r$ is greater than the $\theta$ direction, thus the convergence speed is faster along $r$. Assuming that the optimization along the $r$ direction is mostly complete at this point, then $r$ is approximately equal to the norm of $\boldsymbol{c}$. We can discuss Eq.\ref{eqa:L_theta} as follows:

If the norm of the OCC optimization objective $\boldsymbol{c}$ is set to be too large, $r$ will also be large. To continue optimizing ${L}$ along the direction, the model must rapidly decrease the angle between $\boldsymbol{z_{i}}$ and $\boldsymbol{z_{j}}$, $\forall {i,j}$, to achieve convergence in the $\theta_{{ij}}$ direction. This is harmful because the model rapidly loses sensitivity to the angles of the samples, especially when abnormal samples similar to normal samples are input into the model (as their angle with normal samples is small). The experimental results in Section \ref{sec:impact_norm} confirm this point: as the norm of the OCC optimization objective $\boldsymbol{c}$ increases, the cosine similarity between the abnormal and normal samples increases rapidly, accompanied by a significant decrease in detection performance. Thus the norm of OCC optimization objective is the critical factor affecting performance.

\textbf{Our Assertions.} Based on the above theoretical analysis, we assert that the optimization objective of OCC can be arbitrarily set as long as it adheres to the following constraints:

\textbf{Assertion 1: } The hypersphere center $\boldsymbol{c}$ can be any hypersphere with a suitable norm and should not be large, without the need for prior knowledge about the training dataset. However, obviously degenerate case like $\boldsymbol{c}=0$, which would lead to trivial solutions by simply setting all network parameters to zero, is excluded.

\textbf{Assertion 2:} While we observe that larger norm values tend to lead to decreased performance, the acceptable norm range varies based on different application scenarios due to differing cosine similarities between positive and negative samples. Overall, norm values and cosine similarity appear to have an inverse relationship. For instance, in tasks like OCC on CIFAR-10 \cite{krizhevsky2009cifar}, which focus on semantic deviations, positive and negative samples belong to different categories and have lower similarity, allowing for a larger acceptable norm range. Conversely, in tasks like anomaly detection and segmentation on MVTec AD \cite{mvtec}, which emphasize perceptual deviations, positive and negative samples belong to the same category and exhibit higher similarities, leading to a smaller acceptable norm range.

\textbf{Assertion 3:} For uniformity, we advocate using random vectors sampled from the standard normal distribution as hypersphere centers, with their L2 norms kept at 1. This setting has demonstrated superiority across multiple datasets.


\textbf{Declaration of Advantages.} These findings form the basis for OCC methods to smoothly shift towards online anomaly detection, enabling cold startups without prior training data. Informed by this theory, the devised method can operate independently of distribution assumptions, exhibit model-agnosticism, and remain data-agnostic. A simple implementation is detailed in Section \ref{sec:sim_imp}.

\begin{figure*}[ht]
 \centering
 \includegraphics[width=0.9\linewidth]{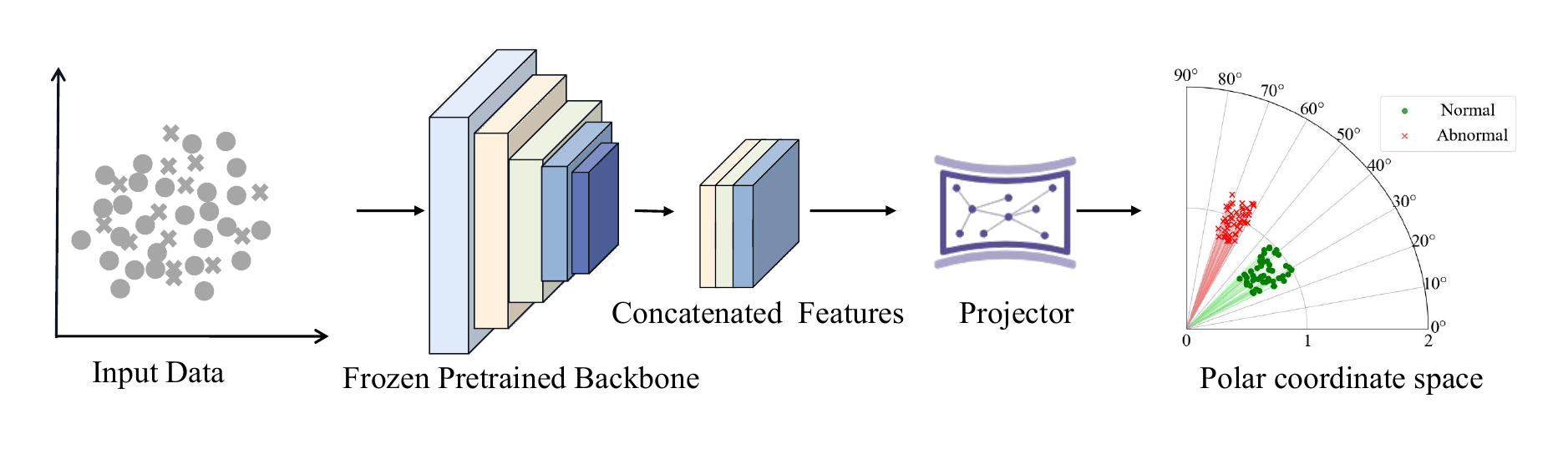}
 \caption{A simple implementation of our method for image anomaly detection and segmentation tasks. We use multi-scale concatenated features, as done in previous methods. As for OCC tasks, the only difference is that we use the penultimate layer of features of the pre-trained backbone. Our method gains excellent performance in polar coordinate space for distinguishing positive and negative samples by setting the optimization objective with suitable norms. }
 \label{fig:method}
\end{figure*}

\subsection{A Simple Implementation of Network}\label{sec:sim_imp}

\textbf{Initialization Phase.} As illustrated in Fig.\ref{fig:method}, the network consists of an encoder pre-trained on the ImageNet dataset and a trainable projector. We follow the standard and common settings, consistent with previous methods: for image one-class classification tasks, the penultimate layer of features from the pre-trained backbone is used. For image anomaly detection and segmentation tasks, multi-scale features are aggregated. The projector is used to map pre-trained features into the hypersphere space centered around $\boldsymbol{c}$. The projector is set as a $1\times 1$ convolutional layer by default. Despite simple, it demonstrates good detection performance. The hypersphere center $\boldsymbol{c}$ is randomly sampled from the standard normal distribution and normalized.

\textbf{Training Phase.} During training, the model first extracts and concatenates the pre-trained feature, $\boldsymbol{f_i} = {E}(\boldsymbol{x_i})$, from normal samples, followed by projecting $\boldsymbol{f_i}$ to the hypersphere: $\boldsymbol{z_i} = \mathcal{P}(\boldsymbol{f_i})$. The loss function is defined using the Euclidean L2 distance to map the pre-trained feature $\boldsymbol{f_i}$ of normal samples into the hypersphere centered around $\boldsymbol{c}$:
\begin{equation}
\fontsize{9pt}{\baselineskip}\selectfont 
{L}_{{W}}=\max \left(\|\boldsymbol{{z_i}}-\boldsymbol{c}\|^ 2-{R}^2,0\right)
\end{equation}

\textbf{Testing Phase.} In the testing phase, for a test sample $\boldsymbol{x}\in \mathcal{X}_{test}$, we compute the pre-trained feature $\boldsymbol{f_{test} } =E(\boldsymbol{x_{test}})$ and the feature after projection $\boldsymbol{z_{test}}=\mathcal{P}(\boldsymbol{f_{test}})$. The decision function for anomaly detection is defined as:
\begin{equation}
\fontsize{9pt}{\baselineskip}\selectfont 
\mathcal{D}\left(\boldsymbol{z_{test}}\right)=\mathbf{1}\{{d}>{R}\}
\end{equation}
where ${d}=\left\|\boldsymbol{z_{test}}-\boldsymbol{c}\right\|$ is the L2 distance between $\boldsymbol{z_{test}}$ and the OCC optimization objective $\boldsymbol{c}$. Additionally, for image anomaly segmentation tasks, the abnormal scores are computed based on the L2 distance to the hypersphere center:
\begin{equation}
\fontsize{9pt}{\baselineskip}\selectfont 
{S}=\max \left(\|\boldsymbol{{z_i}}-\boldsymbol{c}\|^ 2-{R}^2,0\right)
\end{equation}
The pseudo-codes of our method are provided in Supplementary Material.



\section{Evaluation of Methods}
In this section, comprehensive evaluations are conducted to obtain the performance of our method among different tasks: (1) image one-class classification task (Section \ref{sec:i_occ}); (2) industrial image anomaly detection and segmentation task (Section \ref{sec:i_ads}) ; (3) online industrial image anomaly detection and segmentation task (Section \ref{sec:online_ad}). More details can be found in the Supplementary Material.

\subsection{Implementation Details}
\textbf{Evaluation Protocols, Datasets and Metrics.} Our method follows two standard and one self-defined protocols for evaluation. The images in the training and inference phases are first resized to $256\times256$, then center-cropped to $224\times224$, and finally fed into the network after normalization. Any additional transformations are not used. Our code was run on a single NVIDIA 3090 GPU with pytorch 1.10.

\textbf{Protocol A: I-OCC Task.} The anomaly detection performance on the image one-class classification task (I-OCC) are evaluated with five common standard datasets: Fashion MNIST \cite{xiao2017fashion}, CIFAR10 \cite{krizhevsky2009cifar}, CIFAR100 \cite{krizhevsky2009cifar}, CatsVsDogs \cite{elson2007catsvsdogs} and MVTec AD \cite{mvtec}. Except for MVTec AD, the other datasets select one category as normal and all the remaining categories as anomalies.The MVTec AD dataset contains normal data and explicit anomalies in the same category. We evaluate the image-level average ROC AUC \% and the model trainable parameter size.

\textbf{Protocol B: I-ADS Task.} We evaluate methods on the more fine-grained industrial image anomaly detection and segmentation task (I-ADS). Following the standard protocol, the image-level average ROC AUC \% and the pixel-level average ROC AUC \% are evaluated. Besides MVTec AD, two common standard industrial image datasets, MPDD \cite{mpdd} and VisA \cite{visa}, are also used.

\textbf{Protocol C: Online Anomaly Detection.} Our method is evaluated on online industrial image anomaly detection and segmentation task to validate its excellent scalability for online detection, where the batch size is set to 1 for simulating streaming data with MVTec AD, MPDD and VisA datasets. The image-level ROC AUC \% and the pixel-level average ROC AUC \% are evaluated.

\textbf{Methods of Comparison.} We compared our method to the SOTA methods separately under each protocol. This is due to the fact that many SOTA methods in Protocol A have not been evaluated in the I-ADS task of Protocol B, or they are significantly far behind the SOTA. For Protocol A, we compare with the SOTA methods including Deep SVDD \cite{ruff2018deep}, PANDA \cite{panda}, MS \cite{reiss2023mean}, MRot \cite{hendrycks2019MRot}, and IGD \cite{chen2022deep}. For Protocol B, we compare with Patch SVDD \cite{Patchsvdd}, PaDiM \cite{padim}, DRAEM \cite{draem}, FastFlow \cite{fastflow}, STPM \cite{zolfaghari2022unsupervised} and CFA \cite{cfa}. If the original results of these SOTA methods are available, we use them directly, otherwise we perform our own evaluation (if possible). Notably, if other methods reported performance for more than one variant, we defaulted to comparing with the best variant.

\subsection{Protocol A: Evaluation on Image One-Class Classification Task}\label{sec:i_occ}
As depicted in Table \ref{tab:ProtocolA}, our method demonstrates performance on par with, if not surpassing, the current state-of-the-art methods across multiple datasets. Notably, on the MVTec AD dataset, our method achieves a substantial performance gain, surpassing the second-ranked IGD by 5.6\%. It is imperative to highlight that the trainable parameters within our method are minimal among pre-trained-based methods. Our approach is free to select any data-agnostic feature space, thereby obviating the need for pre-collecting datasets. Given the practical constraints wherein datasets often emerge in an incomplete state at the initial stage, this highlights the flexibility, generality, and practicality of our method.


\begin{table}[t]
\centering
\resizebox{.95\columnwidth}{!}{
\begin{tabular}{lcccccc}
\toprule
Methods & DSVDD & PANDA & MS & IGD & MRot & Ours \\
\midrule
FMNIST & 84.8 &\textbf{95.6} & - & 94.4 & 93.2 &\textbf{95.4} \\
CIFAR10 & 64.8 &\textbf{96.2} &\textbf{97.2} & 91.3 & 90.1 & 93.9 \\
CIFAR100 & 67.0 & 94.1 &\textbf{96.4} & 84.3 & 80.1 &\textbf{96.5} \\
CatsVSDogs & 50.5 & 97.3 &\textbf{99.3} & - & 86.0 &\textbf{97.5} \\
MVTec AD & 77.9 & 86.5 & 87.2 &\textbf{93.4} & 65.5 &\textbf{99.0} \\
\midrule
Train Params [MB] & \textbf{1.3} & 55.6 & 55.6 & 28.7 & 11.3 & \textbf{3.7} \\
\bottomrule
\end{tabular}
}
\caption{Protocol A: One-Class Classification Performance (Average Image ROC AUC \%). Bold is the SOTA performance and second ranks. Our method achieves SOTA performance with fewer trainable parameters.}
\label{tab:ProtocolA}
\end{table}

\subsection{Protocol B: Evaluation on Industrial Image Anomaly Detection and Segmentation}\label{sec:i_ads}

In the fine-grained industrial image anomaly detection and segmentation task, we conduct comprehensive comparisons with a multitude of state-of-the-art (SOTA) methods. Moreover, the top-performing OCC method IGD from Protocol A is also benchmarked. As depicted in the Table \ref{tab:ProtocolB}, our method is SOTA on both the image-level average ROC AUC \% and the pixel-level average ROC AUC \%. Notably, while other methods might exhibit high accuracy in a single task (I-OCC/I-ADS), our method achieves SOTA performance in both tasks without requiring elaborate adjustments. In other word, it is proficient in handling both OCC tasks that emphasize semantic deviations and anomaly detection and segmentation tasks that focus on finer-grained statistical deviations. This can be attributed to the remarkably flexible, generic, and data-agnostic optimization objective of our method. 



\begin{table}[t]
\centering
\resizebox{.95\columnwidth}{!}{
\begin{tabular}{lcccccc}
\toprule
Metric & \multicolumn{3}{c}{Average Image ROC AUC \%} & \multicolumn{3}{c}{Average Pixel ROC AUC \%} \\
\midrule
Methods & MVTec AD & MPDD & VisA & MVTec AD & MPDD & VisA \\
\midrule
IGD & 93.4 & 83.1 & 84.7 & 93.1 & 86.8 & 85.4 \\
CFA & 98.1 & 92.3 & 92.0 & 97.1 & 94.8 & 84.3 \\
PaDiM & 90.8 & 70.6 & 89.1 & 96.6 & 95.5 & 98.1 \\
DRAEM & 98.1 & \textbf{94.1} & 88.7 & 97.5 & 91.8 & 93.5 \\
FastFlow & 90.5 & 88.7 & 87.4 & 95.5 & 80.8 & 91.1 \\
STPM & 92.4 & 87.6 & 83.3 & 95.4 & 98.1 & 83.4 \\
PatchSVDD & 92.1 & 86.3 & - & 95.7 & 81.4 & - \\
\midrule
Ours & \textbf{99.0} & 92.3 & \textbf{93.7} & \textbf{98.2} & \textbf{98.2} & \textbf{98.2} \\
\bottomrule
\end{tabular}
}
\caption{Protocol B: Industrial Image Anomaly Detection and Segmentation Performance. Our method outperforms in MVTec AD, MPDD and VisA datasets.}
\label{tab:ProtocolB}
\end{table}

\subsection{Protocol C: Expansion to Online Anomaly Detection Task}\label{sec:online_ad}


As shown in the Table \ref{tab:ProtocolC}, our approach's online detection performance on the MVTec AD, MPDD and VisA datasets is on par with some SOTA methods from Table \ref{tab:ProtocolB}. This performance is remarkably surprising, as it is an unfair comparison for our method: while other methods can repeatedly use pre-collected training samples, ours can only leverage samples in a one-pass manner. This solidly confirms one of the key conclusions of this study: by considering any space with an appropriate norm as the optimization objective for OCC methods, we address the data-agnostic requirement of model initialization in online anomaly detection, eliminating the pre-collection of data. Thus, our method provides a flexible, general, and effective foundation for pushing current offline learning OCC methods towards the online paradigm.

\begin{table}[t]
\centering
\resizebox{\columnwidth}{!}{
\begin{tabular}{lcccccc}
\toprule
Metric & MVTec AD & MPDD & VisA \\
\midrule
Average Image ROC AUC \% & 93.4 & 84.5 & 93.0\\
Average Pixel ROC AUC \% & 95.6 & 97.2 & 98.1 \\
\bottomrule
\end{tabular}
}
\caption{Protocol C: Online Image Anomaly Detection and Segmentation Performance. Our method even outperforms some offline learning SOTA methods in Protocol B.}
\label{tab:ProtocolC}
\end{table}

\section{Ablations}
Thorough ablation experiments and visualization analyses are implemented to evaluate the following topics: (1) visual analysis (Section \ref{sec:vis}) to validate the conclusions of Section \ref{sec:theory_opt_obj} ; (2) the impact of distribution type (Section \ref{sec:impact_dis_type}); (3) the norm of the optimization objective and guidelines for feasible domain of norm (Section \ref{sec:impact_norm}). More ablations can be found in Supplementary Material.

\begin{figure*}[ht]
 \centering
 \includegraphics[width=0.95\linewidth]{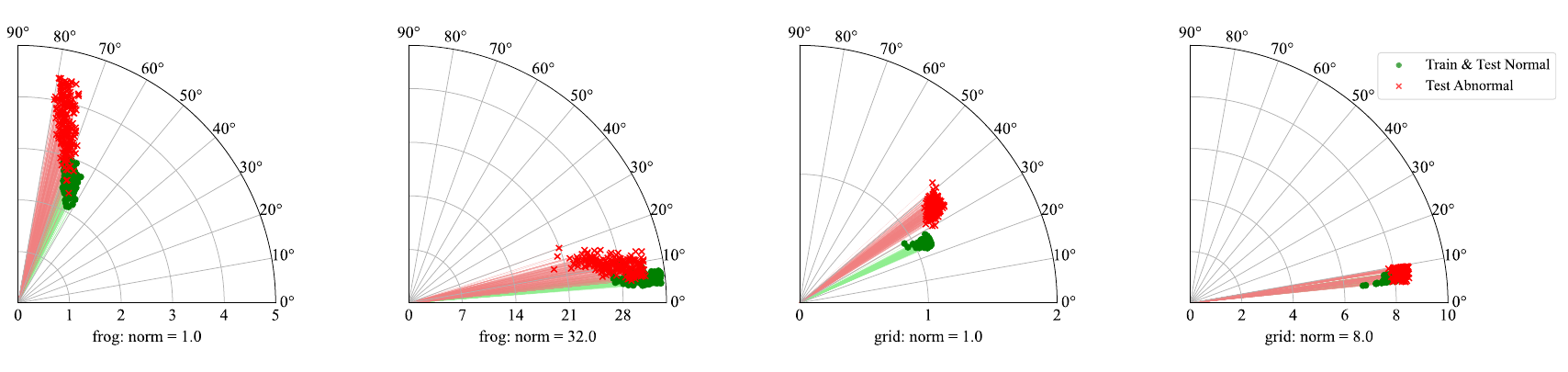}
 \caption{Illustration of the model's positive and negative sample discrimination performance for the \textbf{frog} category in CIFAR10 and the \textbf{grid} category in the MVTec AD dataset. Visualizing norm-based and angular discrimination. With the norm value increases, OCC model gradually diminishes sensitivity to angular differences between positive and negative samples. For instance, in the grid category, the angle between negative and positive samples decreases significantly at norm 8 compared to norm 1, implying increased challenge in sample discrimination.}
 \label{fig:ablation_angular_sensitivity}
\end{figure*}

\subsection{Visualization}\label{sec:vis}

\textbf{Norm-based Degradation of OCC methods.} By visualizing in Fig. \ref{fig:ablation_angular_sensitivity}, we examine the positive and negative sample discriminatory power of the OCC method with the optimization objective set to different norms. Observe Fig. \ref{fig:ablation_angular_sensitivity}, the conclusion of our theoretical analysis in Section 3.2 is proved: when the norm of the OCC optimization objective is too large, the neural network gradually loses its angular sensitivity to positive and negative samples, which finally degrades the detection performance.

\subsection{Impact of Distribution Type}\label{sec:impact_dis_type}
The distribution types of hypersphere centers are first analyzed in Table \ref{tab:Ablations_Distribution}. When the norm of hypersphere centers is strictly limited to 1, the detection accuracy rarely decreases, regardless of whether the hypersphere centers come from pre-trained features or from random distributions. This validates assertion 1 in Section \ref{sec:theory_opt_obj}: when the norms are strictly consistent, the optimization objective of the OCC method can be set to data-agnostic any space, and the average of the pre-trained features commonly used by the previous method is just one of the solutions.

\begin{table}[t]
\centering
\resizebox{0.95\columnwidth}{!}{
\begin{tabular}{lccccc}
\toprule
Distribution Type & L2 Normalization & CIFAR10 & MVTec AD \\
\midrule
Pre. Features & \checkmark & 93.88 & 98.97 \\
$\mathcal{N}(0,1)$ & $\checkmark$ & 93.93 & 99.00 \\
$\mathcal{U}(0,1)$ & $\checkmark$ & 93.91 & 98.96 \\
$\mathcal{J}(1)$ & $\checkmark$ & 93.86 & 98.93 \\
\bottomrule
\end{tabular}
}
\caption{Impact of distribution type of hypersphere center on anomaly detection performance. Average Image ROC AUC \% is reported. Distribution type includes pre-trained features average, standard normal distribution $\mathcal{N}(0,1)$, uniform distribution $\mathcal{U}(0,1)$ and $\mathcal{J}(1)$ with each element is 1.}
\label{tab:Ablations_Distribution}
\end{table}

\subsection{Impact of Norm of Hypersphere Center}\label{sec:impact_norm}
\begin{figure}[ht]
 \centering
 \includegraphics[width=0.8\linewidth]{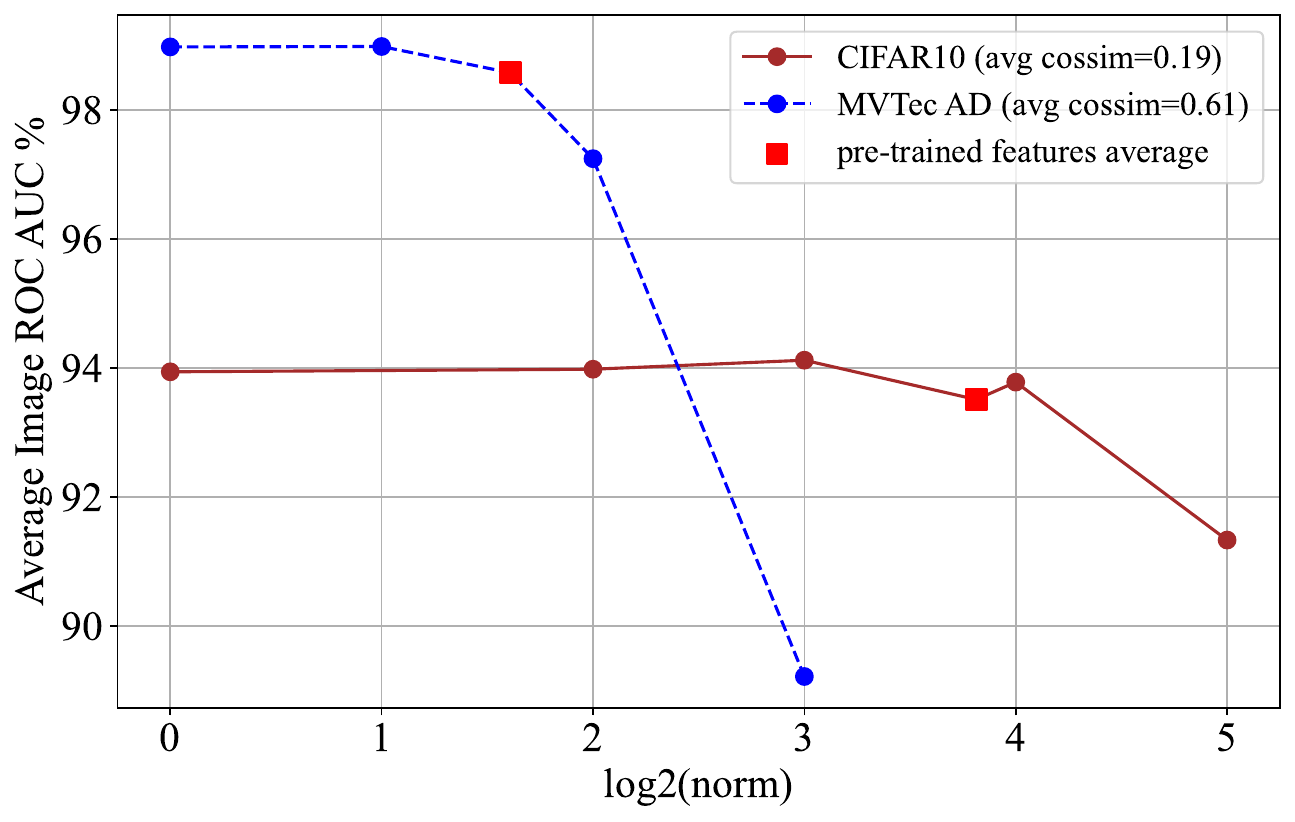}
 \caption{Impact of the OCC optimization objective's norm.}
 \label{fig:ablation_norm}
\end{figure}

\textbf{Impact of the norm of the OCC optimization objective.} As depicted in Fig. \ref{fig:ablation_norm}, with the hypersphere center’s norm increases, the detection performance on different datasets show a similar decreasing trend. This observation suggests that the optimization objective’s norm of the OCC method has a notable impact on the detection performance. Too large norms would weaken the ability of the OCC method to discriminate between positive and negative samples. But the different datasets have varying sensitivity thresholds.

\begin{figure}[ht]
 \centering
 \includegraphics[width=0.8\linewidth]{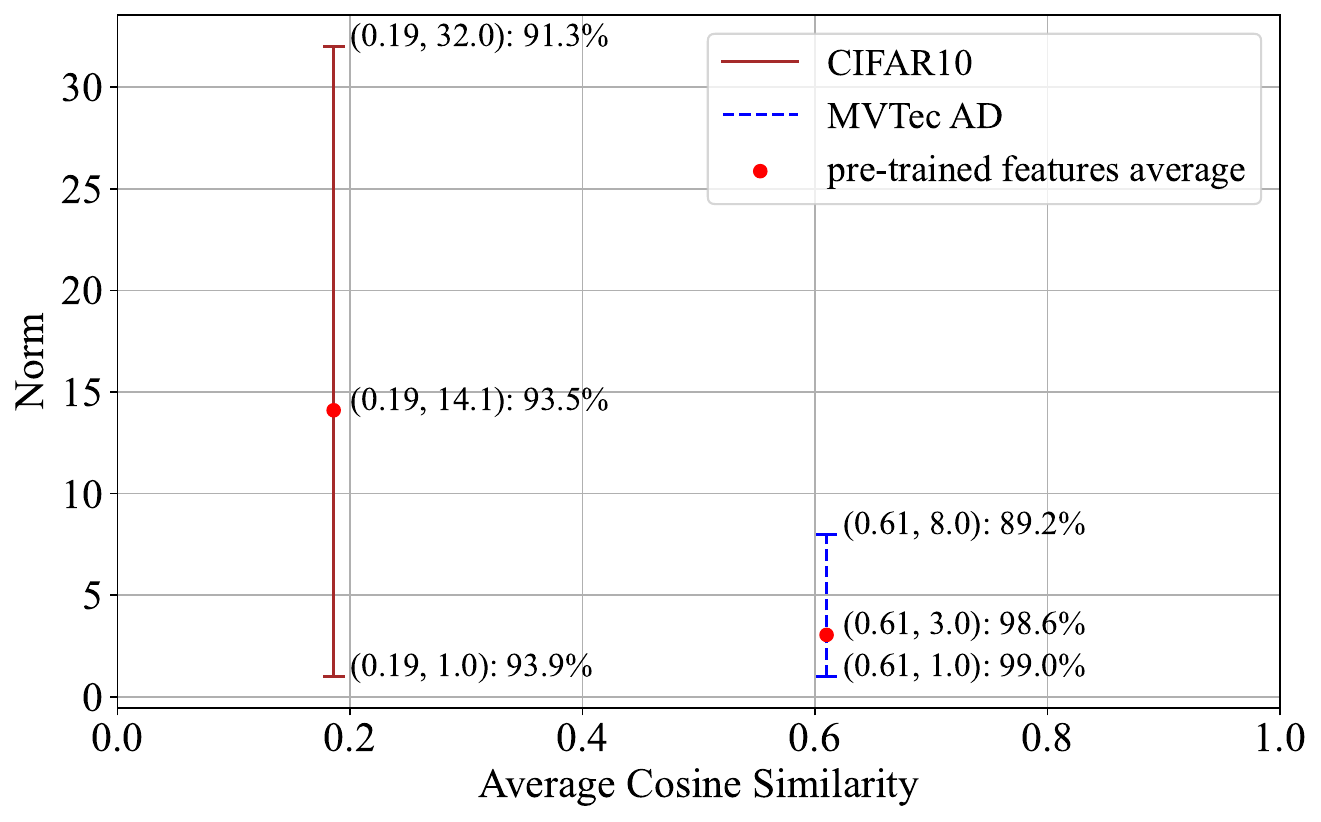}
 \caption{Feasible domains of the optimization objective's norm. Average cosine similarity between positive and negative samples, norm value, average image ROC AUC \% are reported.}
 \label{fig:ablation_feasible_domain}
\end{figure}

\textbf{Guidelines for determining the feasible domain of the optimization objective.}
As shown in Fig. \ref{fig:ablation_norm}, we observe that the performance degradation is significantly different across datasets: the method is more robust to norm changes of the hypersphere center on the CIFAR10 dataset, while it exhibits more sensitivity on the MVTec AD dataset. We attribute this phenomenon to the semantic similarity between the positive and negative categories of the dataset. This validates our assertion 2 in Section \ref{sec:theory_opt_obj}: datasets with higher similarity between positive and negative samples entail a narrower feasible domain for the optimization objective's norm of OCC.

To determine the feasible domain of the OOC optimization objective norm on different datasets, we have established empirical guiding principles based on experimental results. As shown in Fig. \ref{fig:ablation_feasible_domain}, the pre-trained backbone is used to extract image features, and then the average cosine similarity of the pre-trained features for positive and negative samples is calculated (0.19 for CIFAR10 and 0.61 for MVTec AD). When striving for an OCC method detection accuracy surpassing 90.0\%, the feasible domain for the optimization objective's norm is identified as [1,32] for CIFAR10 and [1,8) for MVTec AD datasets. Significantly, it is noteworthy that the optimization objective configuration strategy employed by previous OCC methods (i.e., the average of pre-trained features) constitutes a solution within the feasible domains that we have given. Readers are encouraged to calculate the average cosine similarity for their own datasets, subsequently opting for an appropriate norm for the optimization objective of OCC based on the insights derived from the test results illustrated in Fig. \ref{fig:ablation_feasible_domain}.



\section{Conclusion}
In this study, we conducted a rigorous investigation into the optimization objective of one-class classification (OCC) and unveiled its interpretation in polar coordinates. The key to distinguishing positive and negative samples lies in the careful selection of the radius (norm) value, ensuring the model's sensitivity and discriminability among angular. We established the guidelines for determining feasible norm ranges of OCC optimization objectives. Furthermore, these insights paved the way for a simple, data-agnostic deep one-class classification method, facilitating OCC's transition towards online anomaly detection. Extensive experiments confirm the effectiveness of our theory and model, resulting in state-of-the-art performance across OCC tasks and industrial image anomaly detection and segmentation.

\textbf{Limitations.} However, it's important to note that our method considers cases with entirely normal training sets and does not address potential anomalies, particularly relevant in online scenarios. Our initial exploration into online anomaly detection offers room for further enhancements. Future work will refine our theory and model to achieve complete online anomaly detection capabilities.
\bibliography{arxiv}

\appendix
\section*{\centering Appendix}
\addcontentsline{toc}{section}{Appendix}

\setcounter{figure}{0}
\setcounter{table}{0}
\renewcommand{\thefigure}{S\arabic{figure}}
\renewcommand{\thetable}{S\arabic{table}}

\section{Datasets}
\begin{figure*}[ht]
\centering
\includegraphics[width=\linewidth]{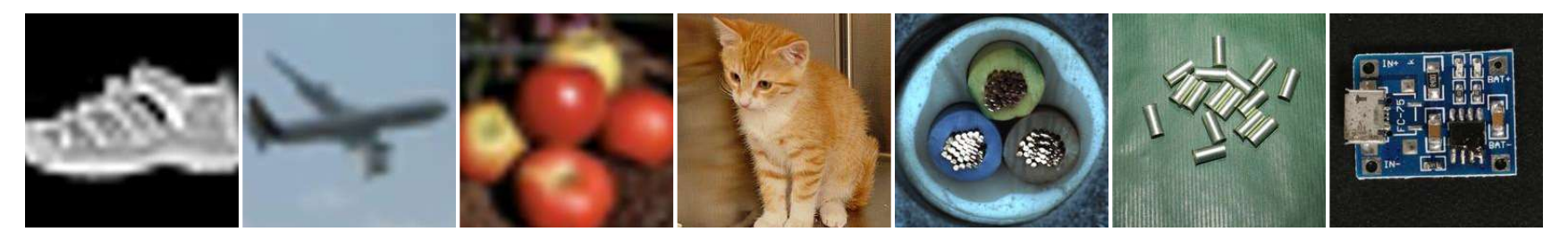}
\includegraphics[width=\linewidth]{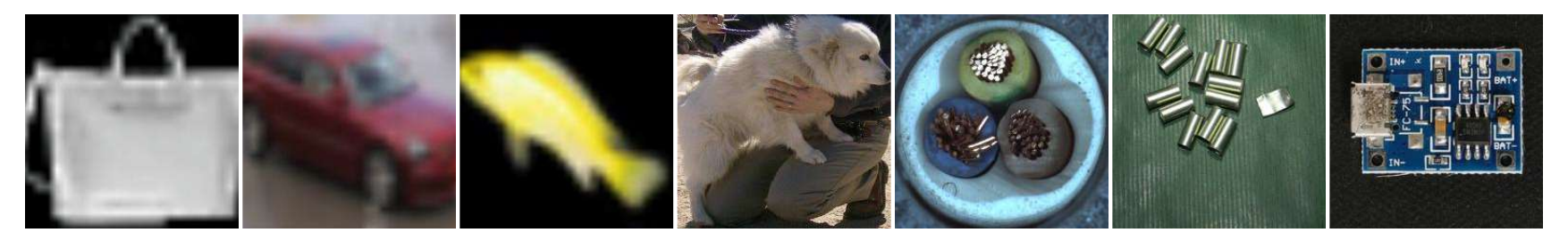}
\caption{Samples of datasets. From left to right are Fashion MNIST, CIFAR10, CIFAR100, CatsVsDogs, MVTec AD, MPDD and VisA. The first row is the normal samples and the second row is the abnormal samples. Following the standard settings, for the Fashion MNIST, CIFAR10, CIFAR100, and CatsVsDogs datasets, we select any one category in the dataset as the normal category, e.g., airplanes in CIFAR10, and treat the other categories as the anomalous categories. For MVTec AD, MPDD and VisA, positive and negative samples are in the same category, e.g., bent wire type anomalies of cable category in MVTec AD dataset.}
\label{fig:Datasets}
\end{figure*}

In this paper, the OCC methods are evaluated on seven datasets, including Fashion MNIST, CIFAR10, CIFAR100, CatsVsDogs, MVTec AD, MPDD, and VisA. The category information is shown in Table \ref{tab:Datasets} and some of the samples are shown in Fig. \ref{fig:Datasets}. For the Fashion MNIST, CIFAR10, CIFAR100, and CatsVsDog datasets, we selected one category as the normal category and the others as the abnormal categories. Meanwhile, for the MVTec AD, MPDD and VisA datasets, normal and abnormal samples belong to the same category.

\section{Methods Details}

\subsection{Network Structure and Optimizer} Since we proved that the optimization objective of the OCC method can be set to arbitrary space with a suitable norm (in Section 3.2), in all evaluations, the hypersphere center is by default random sampled from the standard normal distribution $\mathcal{N}(0,1)$ with L2 normalization. The random seed is fixed to 1024. In \textbf{Protocol A}, the batch size is set to 64. We first use the penultimate layer of the pre-trained ResNet152 backbone on the ImageNet dataset as the embedded feature $f$ (with dimension 2048). The projector is set by default to be a $1\times1$ convolutional layer with a bias term, which maps the feature $f$ into the latent feature $z$, keeping the feature dimension unchanged. The optimizer is set to Adam, with an initial learning rate of 1e-4 and a decay rate of 5e-4. Training epochs are 20. In \textbf{Protocol B}, the batch sizes are set by default to 16, 8, and 8 for MVTec AD, MPDD, and VisA. The embedded features $f \in \mathcal{R}^{B\times1792\times56\times56}$, consist of layers 2, 3, and 4 of the pre-trained Wider Resnet 50 backbone, which is a common and standard setup for industrial image anomaly detection and segmentation. The optimizer settings are the same as in Protocol A. Training epochs are 50. In \textbf{Protocol C}, all settings are the same as in Protocol B, except that the batch size is 1 to simulate the online streaming data, where the sample is one-by-one input.

\begin{table}[t]
\centering
\resizebox{\columnwidth}{!}{
\begin{tabular}{lcccc}
\toprule
Datasets & Category & Train & Test & Image Size (min/max)\\
\midrule
FMNIST & 10& 60000 & 10000 & 28/28 \\
CIFAR10 & 10 & 50000 & 10000 & 32/32 \\
CIFAR100 & 100 & 50000 & 10000 & 32/32 \\
CatsVSDogs & 2 & 8750 &7500 & 280/500 \\
MVTec AD & 15 & 4096 & 1258 & 512/512 \\
MPDD & 6 & 1064 & 282 & 1024/1024 \\
VisA & 12 & 10621 & 1200 & 960/1562 \\
\bottomrule
\end{tabular}
}
\caption{Detailed information of datasets, including the number of categories, the total number of samples in the training and test sets, and the minimum/maximum image sizes.}
\label{tab:Datasets}
\end{table}

\subsection{Pseudo Codes}

We provide the pseudo-codes of our method for image one-class classification tasks (Protocol A) and industrial image anomaly detection and segmentation tasks (Protocol B), which are shown in Algorithm \ref{alg:protocol_a} and Algorithm \ref{alg:protocol_b}.

\begin{algorithm}
\caption{Our Method for Image One-Class Classification Tasks.}
\begin{algorithmic}[1]\label{alg:protocol_a}
\STATE \textbf{Input:} ImageNet pre-trained encoder $E$, Projector $\mathcal{P}$ (e.g. $1\times1$ conv), Radius of hypersphere $R=1e^{-5}$.\
\STATE \textbf{Training Stage:}
\STATE \quad Initialization: $c \leftarrow$ standard normal distribution(0, 1)
\STATE \quad Train sample $x_{\text{train}} \leftarrow$ normal samples
\STATE \quad Pre-trained feature $f_{\text{train}} \leftarrow E(x_{\text{train}})$
\STATE \quad  Mapped feature $z_{\text{train}} \leftarrow P(f_{\text{train}})$
\STATE \quad  $L_W \leftarrow \max(||z_{\text{train}} - c||^2 - R^2, 0)$
\STATE \quad  $L_W$.backward()

\STATE \textbf{Testing Stage:}
\STATE \quad Test sample $x_{\text{test}} \leftarrow$ test dataset
\STATE \quad Pre-trained feature $f_{\text{test}} \leftarrow E(x_{\text{test}})$
\STATE \quad  Mapped feature $z_{\text{test}} \leftarrow P(f_{\text{test}})$
\STATE \quad Distance $d \leftarrow ||z_{\text{test}} - c||$
\STATE \textbf{Output:} OCC detection decision $\mathbf{1} \{ d > R \}$
\end{algorithmic}
\end{algorithm}

\begin{algorithm}
\caption{Our Method for Industrial Image Anomaly Detection and Segmentation Tasks.}
\begin{algorithmic}[1]\label{alg:protocol_b}
\STATE \textbf{Input:} ImageNet pre-trained encoder $E$, Projector $\mathcal{P}$ (e.g. $1\times1$ conv), Radius of hypersphere $R=1e^{-5}$.\
\STATE \textbf{Training Stage:}
\STATE \quad Initialization: $c \leftarrow$ standard normal distribution(0, 1)
\STATE \quad Train sample $x_{\text{train}} \leftarrow$ normal samples
\STATE \quad Multi-Scale Pre-trained feature $\{f^{(i)}_{\text{train}}\}_{i=1}^3 \leftarrow E(x_{\text{train}})$
\STATE \quad Concated feature $f_{\text{train}} \leftarrow$ Concated$(\{f^{(i)}_{\text{train}}\}_{i=1}^3)$
\STATE \quad  Mapped feature $z_{\text{train}} \leftarrow P(f_{\text{train}})$
\STATE \quad  $L_W \leftarrow \max(||z_{\text{train}} - c||^2 - R^2, 0)$
\STATE \quad  $L_W$.backward()

\STATE \textbf{Testing Stage:}
\STATE \quad Test sample $x_{\text{test}} \leftarrow$ test dataset
\STATE \quad Pre-trained feature $f_{\text{test}} \leftarrow E(x_{\text{test}})$
\STATE \quad  Mapped feature $z_{\text{test}} \leftarrow P(f_{\text{test}})$
\STATE \textbf{Output:} Anomaly Score $||z_{\text{test}} - c||$
\end{algorithmic}
\end{algorithm}

\section{Evaluation of Methods}
\subsection{Accuracy} 
\begin{figure*}[ht]
\centering
\includegraphics[width=\linewidth]{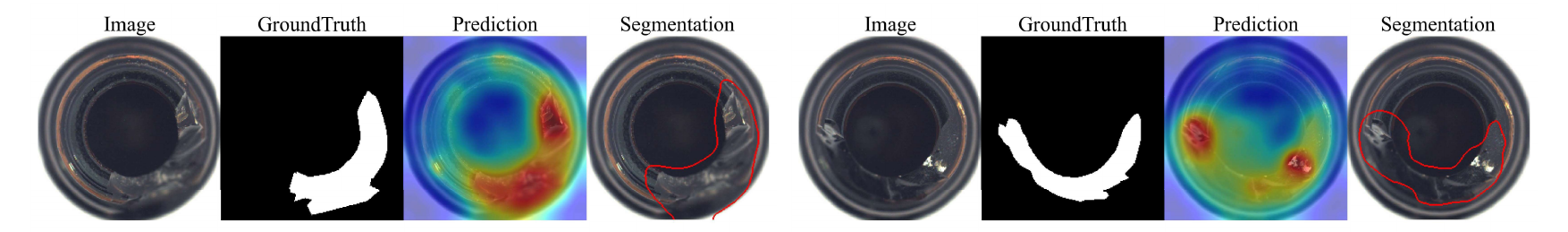}
\includegraphics[width=\linewidth]{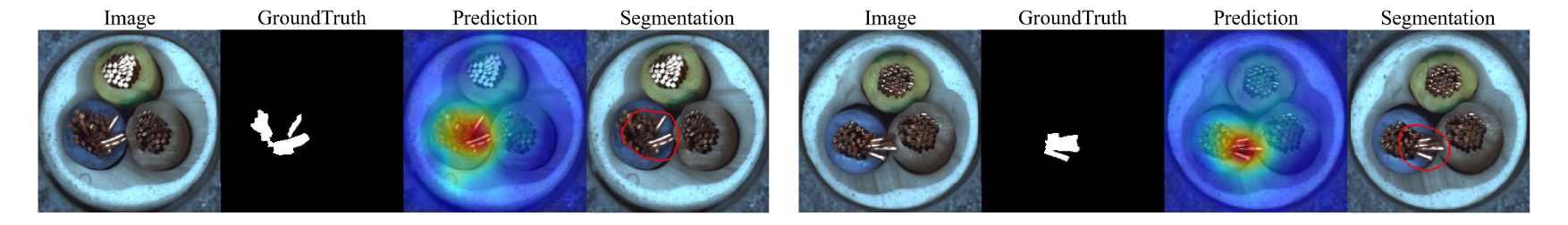}
\includegraphics[width=\linewidth]{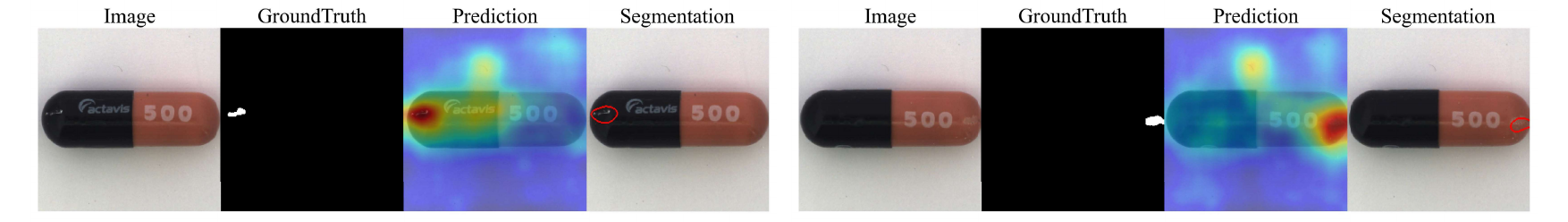}
\includegraphics[width=\linewidth]{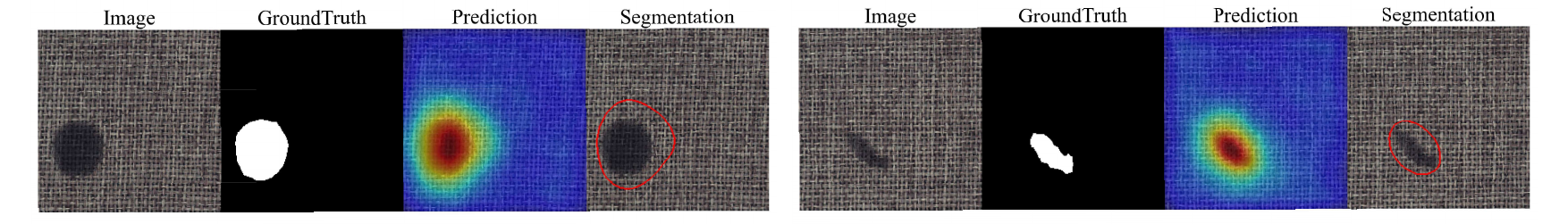}
\includegraphics[width=\linewidth]{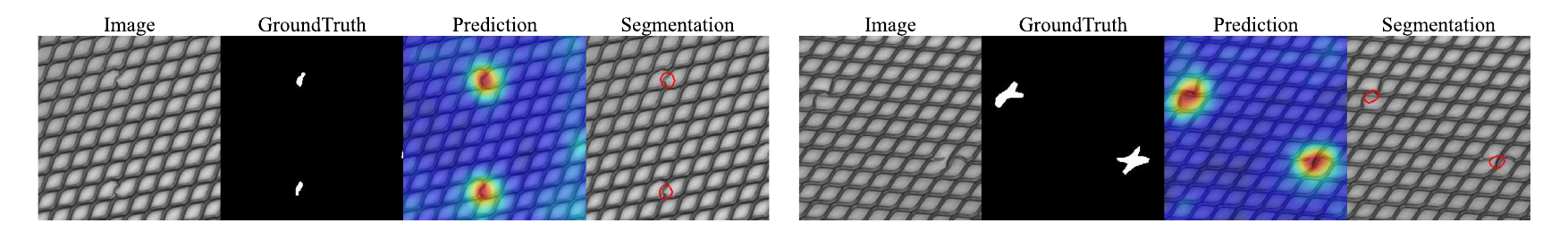}
\includegraphics[width=\linewidth]{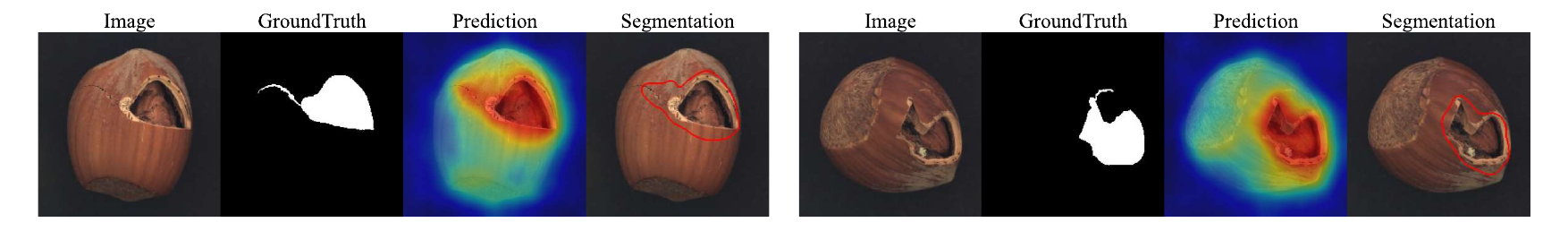}
\includegraphics[width=\linewidth]{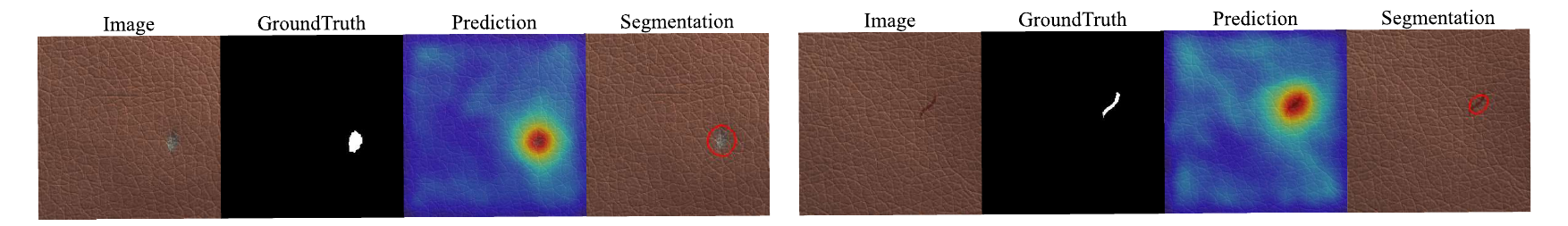}
\includegraphics[width=\linewidth]{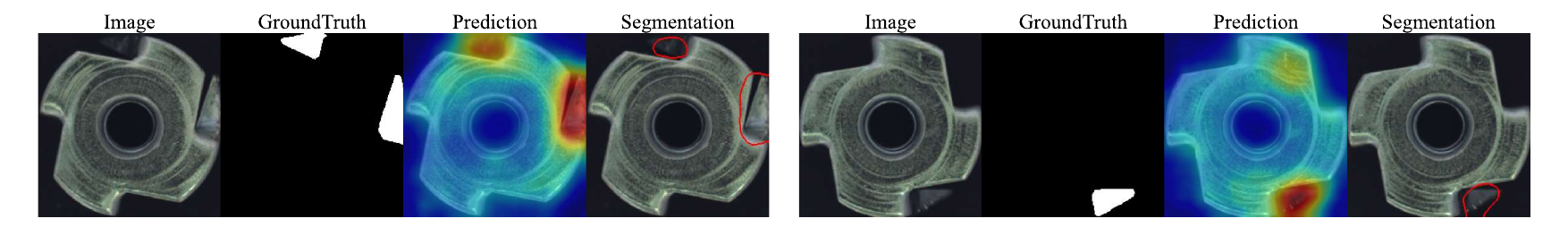}
\includegraphics[width=\linewidth]{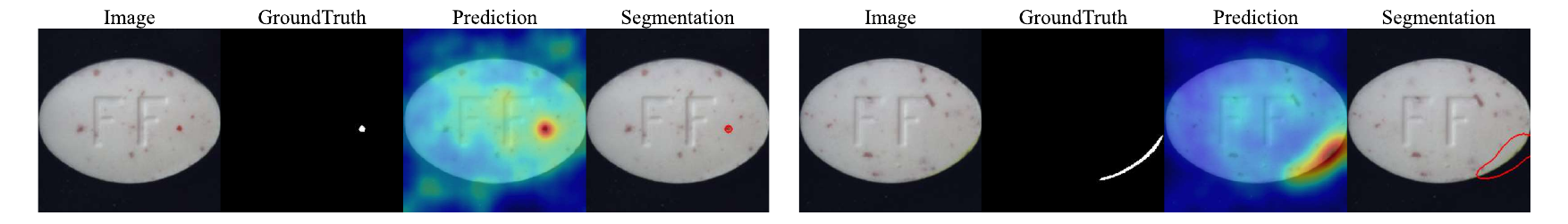}
\caption{ Visualization of 9 categories of MVTec AD: bottle, cable, capsule, carpet, grid, hazelnut, leather, metal nut and pill.}
\label{fig:defect_vis_1}
\end{figure*}

\begin{figure*}[ht]
\centering
\includegraphics[width=\linewidth]{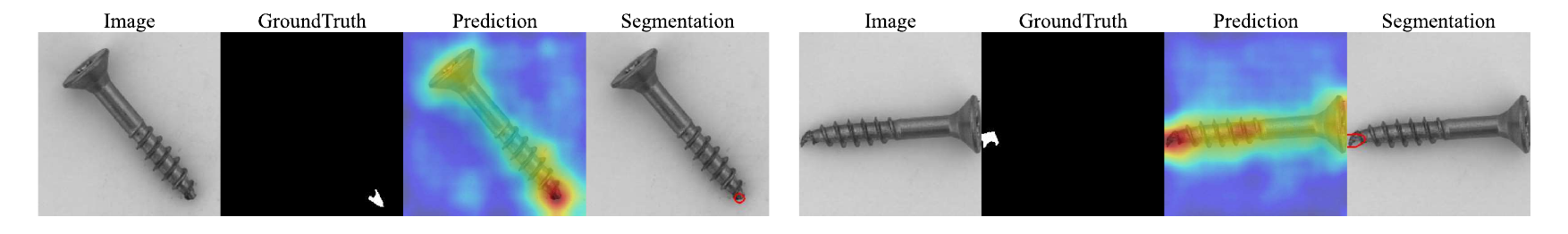}
\includegraphics[width=\linewidth]{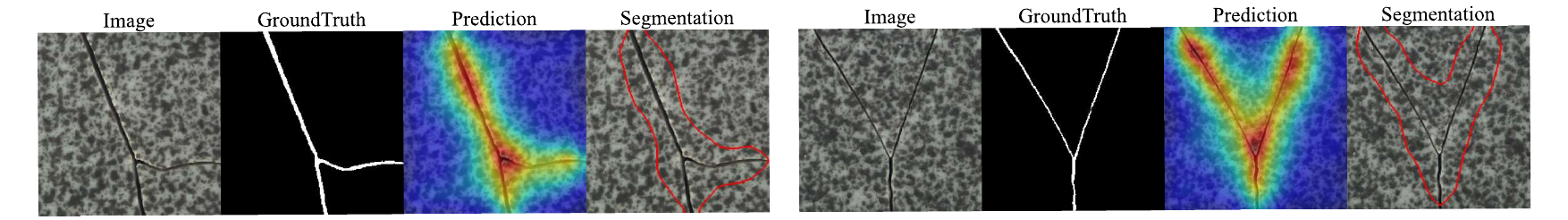}
\includegraphics[width=\linewidth]{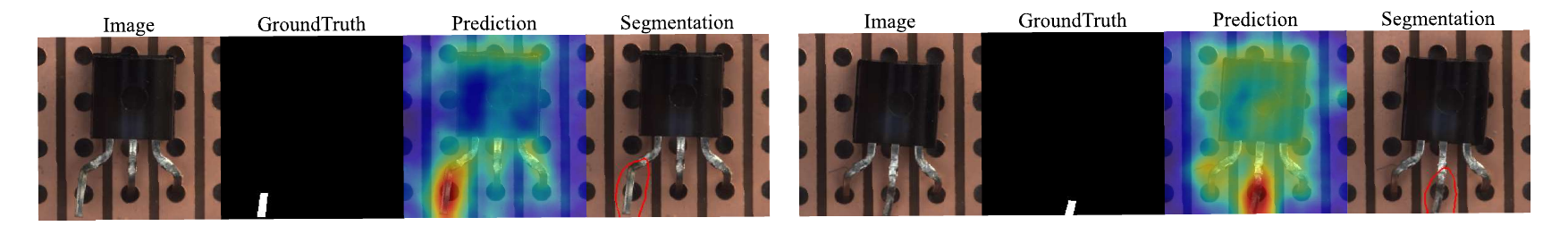}
\includegraphics[width=\linewidth]{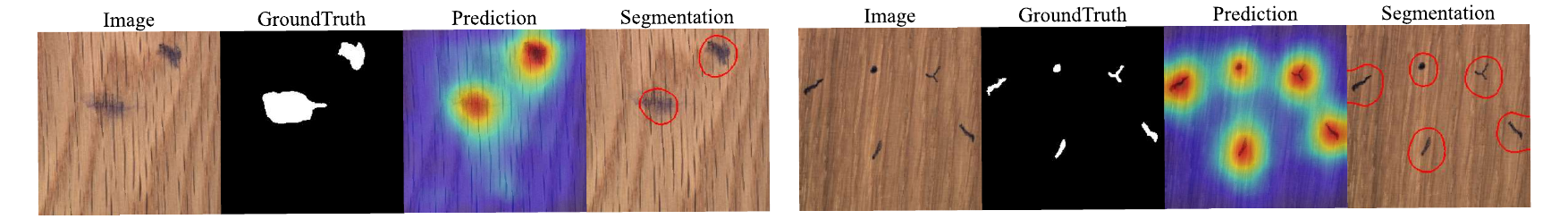}
\includegraphics[width=\linewidth]{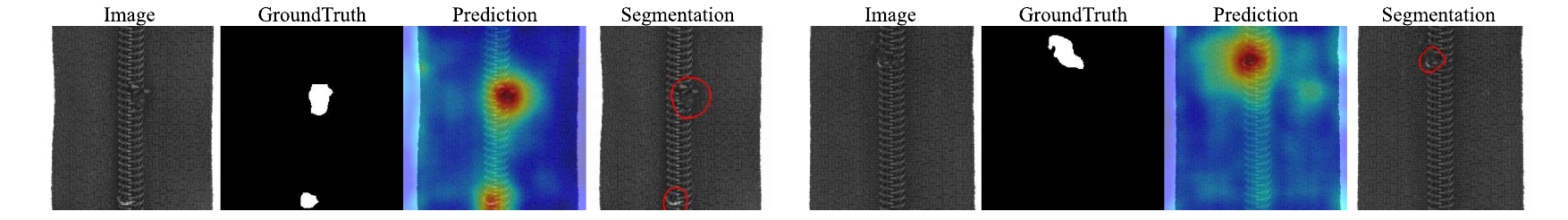}
\caption{ Visualization of 6 categories of MVTec AD: screw, tile, transistor, wood and zipper.}
\label{fig:defect_vis_2}
\end{figure*}

This section shows a detailed comparison of our methods and others for each category under standard protocols. We report the image-level average ROC AUC \% in Table \ref{tab:protocol_b_performance}. Additionally, the visualization of detection results are shown in Fig. \ref{fig:defect_vis_1} and Fig. \ref{fig:defect_vis_2}.

\subsection{Imapct of Network Structure and Projector} We report the ablation of network structure and projector in Table \ref{tab:backbones} and Table \ref{tab:projectors}. Obviously, our method can be well adapted to different backbones and projectors.

\begin{table}[t]
\centering
\begin{tabular}{l|cc}
\toprule
\textbf{Backbones} & CIFAR10 & MVTec AD \\
\midrule
Resnet18 & 88.6 & 78.49\\
\midrule
Wider Resnet50 & 90.9 & 99.0 \\
\midrule
Resnet152 & 93.9 & 97.7 \\
\bottomrule
\end{tabular}
\caption{Ablations of different backbones. The image-level average ROC AUC \% is reported.}
\label{tab:backbones}
\end{table}

\begin{table}[t]
\centering
\begin{tabular}{l|cc}
\toprule
\textbf{Projectors} & CIFAR10 & MVTec AD \\
\midrule
Conv$1 \times 1$ & 93.9 & 99.0\\
\midrule
Linear & 93.9 & 97.3 \\
\bottomrule
\end{tabular}
\caption{Ablations of different projectors. The image-level average ROC AUC \% is reported.}
\label{tab:projectors}
\end{table}

\begin{table*}[t]
\centering
\begin{tabular}{l|ccccccc|cc}
\toprule
\textbf{Method} & IGD & CFA & PaDiM & DRAEM & FastFlow & STPM & Patch SVDD & Ours \\
\midrule
Bottle & 100.0 & 100.0 & 99.4 & 96.9 & 91.2 & 100.0 & 96.9 & 100.0\\
Cable & 85.9 & 99.4 & 85.0 & 93.4 & 99.2 & 74.8 & 89.5 & 98.4\\
Capsule & 83.3 & 99.9 & 94.9 & 96.1 & 87.5 & 94.4 & 76.7 & 96.3\\
Carpet & 80.0 & 94.7 & 88.9 & 96.3 & 88.1 & 92.5 & 95.7 & 99.9\\
Grid & 60.0 & 99.7 & 90.4 & 100.0 & 98.8 & 77.4 & 89.9 & 99.6\\
Hazelnut & 98.2 & 99.6 & 99.1 & 100.0 & 96.1 & 95.6 & 91.2 & 100.0\\
Leather & 90.8 & 98.6 & 84.3 & 100.0 & 90.8 & 95.3 & 75.6 & 100.0\\
Metalnut & 84.8 & 96.5 & 87.6 & 99.4 & 98.0 & 94.0 & 94.0 & 100.0\\
Pill & 76.4 & 98.5 & 98.1 & 96.8 & 97.8 & 93.0 & 78.6 & 97.9\\
Screw & 67.0 & 98.3 & 92.2 & 99.2 & 63.2 & 97.6 & 82.2 & 94.8\\
Tile & 98.3 & 99.9 & 86.5 & 100.0 & 100.0 & 99.0 & 96.8 & 99.6\\
Toothbrush & 90.8 & 100.0 & 100.0 & 99.7 & 88.1 & 100.0 & 98.9 & 100.0\\
Transformer & 90.5 & 99.4 & 94.1 & 94.2 & 90.8 & 97.2 & 89.4 & 99.3\\
Wood & 94.5 & 97.4 & 80.8 & 99.2 & 67.5 & 96.3 & 98.4 & 99.9\\
Zipper & 90.5 & 88.8 & 80.9 & 99.7 & 100.0 & 78.5 & 98.0 & 99.2\\
\midrule
\textbf{Avg} & 86.3 & 98.1 & 90.8 & 98.1 & 90.5 & 92.4 & 90.1 & 99.0 \\
\bottomrule
\end{tabular}
\caption{Industrial image anomaly detection and segmentation performance comparison on MVTec AD. The image-level average ROC AUC \% is reported.}
\label{tab:protocol_b_performance}
\end{table*}

\section{Visualization}

\subsection{Norm-based Degradation of OCC methods}

\begin{figure*}[ht]
\centering
\includegraphics[width=0.95\linewidth]{figures/ablation_angular_sensitivity.pdf}
\includegraphics[width=\linewidth]{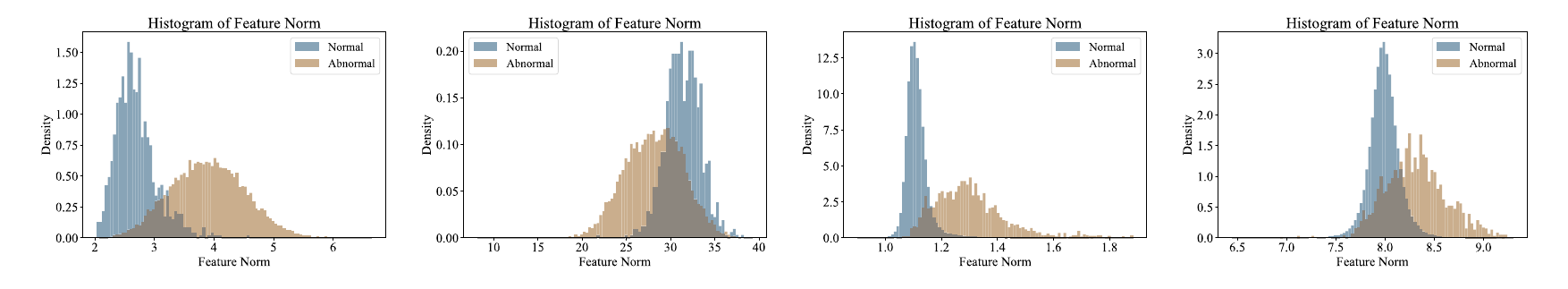}
\includegraphics[width=\linewidth]{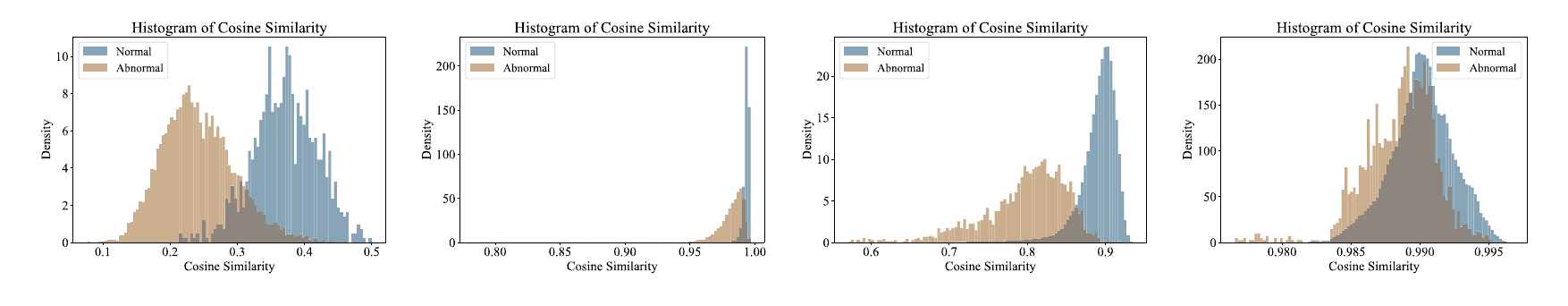}
\caption{Inappropriate norm of OCC's optimization objective can lead to significant degradation of detection performance. Here are two examples for the frog category (first two columns) in CIFAR10 and the grid category (last two columns) in MVTec AD. For the frog category, the image-level average ROC AUC decreases from 95.8\% (norm=1.0) to 92.7\% (norm=32.0). The cosine similarity between positive and negative samples also becomes more indistinguishable at norm=32.0. A similar trend was seen in the grid category of MVTec AD, where the image-level average ROC AUC \% decreased from 99.6\% (norm=1.0) to 75.2\% (norm=8.0).}
\label{fig:ablation_angular}
\end{figure*}

We report in detailed the impact of the norm of the OCC optimization objective on the model's ability to discriminate between positive and negative samples. Fig. \ref{fig:ablation_angular} illustrates the frog category from CIFAR10 and the grid category from MVTec AD. As the norm of the OCC optimization objective increases, the model's ability to distinguish between positive and negative samples on both categories shows a similar significant degradation.

\subsection{Impact of Norm of Hypersphere Center}

\begin{figure*}[ht]
\centering
\includegraphics[width=0.95\linewidth]{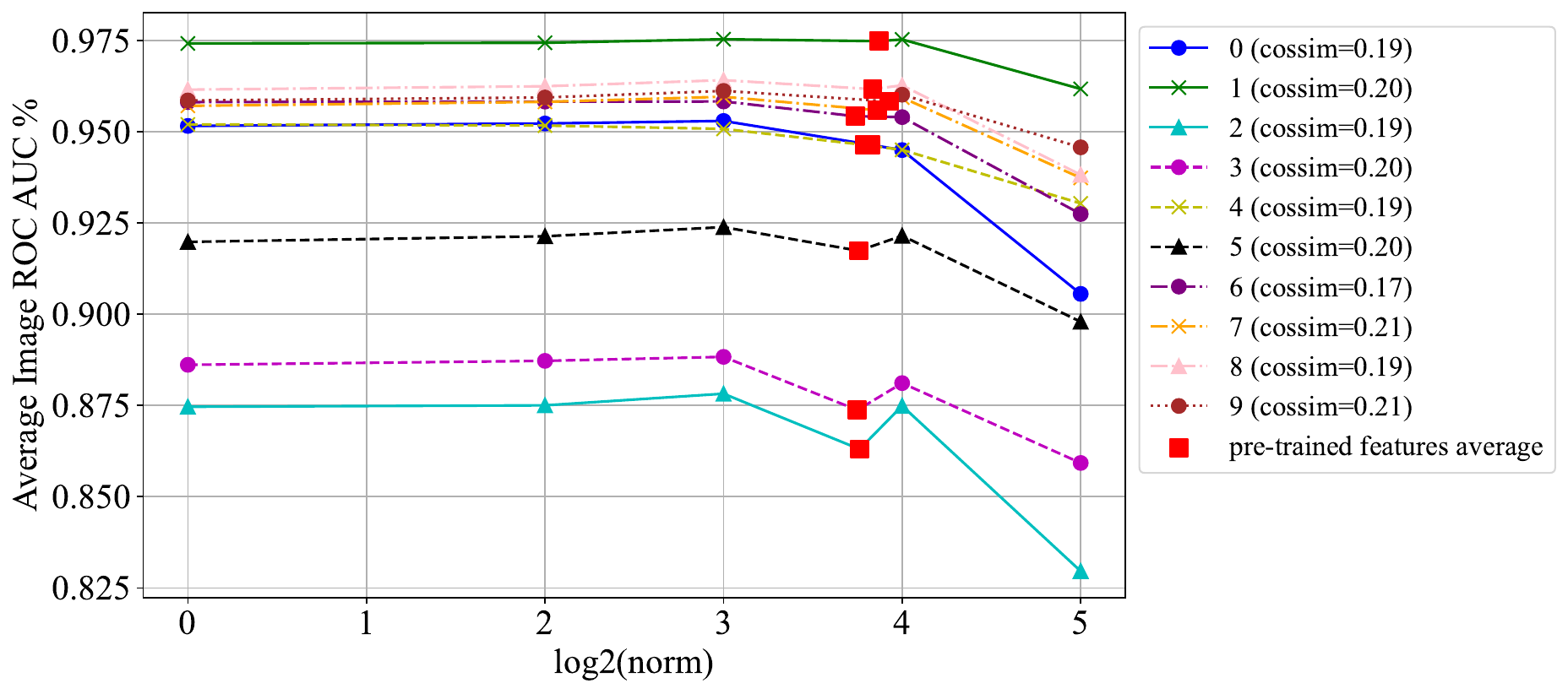}
\caption{Impact of OCC optimization objective's norm on each category in CIFAR10.}
\label{fig:cifar_per_fig}
\end{figure*}

Fig. \ref{fig:cifar_per_fig}, Table \ref{tab:cifar_per_tab} and Table \ref{tab:mvtec_per_tab} show the impact of the different norms of the OCC optimization objective on each category. These results support the conclusions and results in Section 5.3: the feasible domain of the OCC optimization objective norm is closely related to the cosine similarity between positive and negative samples. The higher the cosine similarity, the smaller the feasible domain.

\begin{table}[t]
\centering
\begin{tabular}{l|ccccc}
\toprule
\textbf{Norms} & 1.0 & 4.0 & 8.0 & 16.0 & 32.0 \\
\midrule
Avg & 93.9 & 94.0 & 94.1 & 93.8 & 91.3\\
\bottomrule
\end{tabular}
\caption{Detection performance under different norms of OCC optimization objective on CIFAR10. The image-level average ROC AUC \% is reported.}
\label{tab:cifar_per_tab}
\end{table}

\begin{table}[t]
\centering
\begin{tabular}{l|ccccc}
\toprule
\textbf{Norms} & 1.0 & 2.0 & 4.0 & 8.0\\
\midrule
Avg & 99.0 & 99.0 & 97.2 & 89.2\\
\bottomrule
\end{tabular}
\caption{Detection performance under different norms of OCC optimization objective on MVTec AD. The image-level average ROC AUC \% is reported.}
\label{tab:mvtec_per_tab}
\end{table}

\subsection{Impact of Distribution Type of Hypersphere Center} 

\begin{figure*}[ht]
\centering
\includegraphics[width=0.95\linewidth]{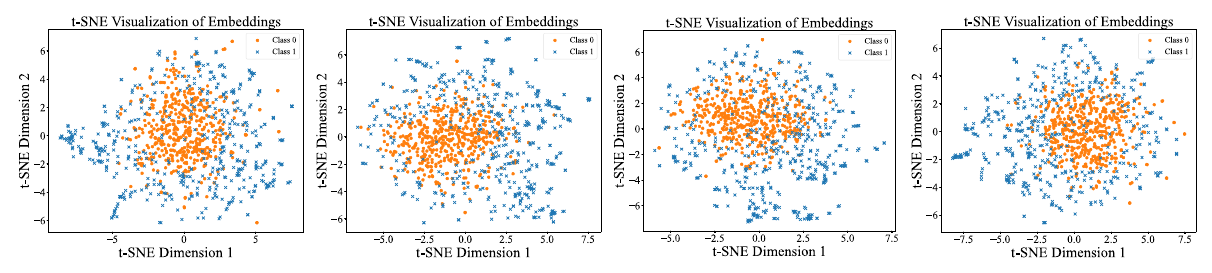}
\caption{Frog category in CIFAR10. From left to right are the t-SNE plots for the case where the OCC optimization objective obeys the pre-trained features average, standard normal distribution $\mathcal{N}(0,1)$, uniform distribution $\mathcal{U}(0,1)$ and $\mathcal{J}(1)$ with each elemnt is 1, respectively.}
\label{fig:t-sne}
\end{figure*}

We presents the t-SNE plot of the output features of the OCC model. As shown in Fig. \ref{fig:t-sne}, the output features of the OCC method exhibit similar properties despite the fact that the normalized optimization objective obeys different distributions, and the normal/abnormal discrimination performance is trivially affected.

\end{document}